\documentclass[conference]{IEEEtran}
\IEEEoverridecommandlockouts
\usepackage{comment}
\usepackage{cite}
\usepackage{amsmath,amssymb,amsfonts}
\usepackage{algorithmic}
\usepackage{graphicx}
\usepackage{textcomp}
\usepackage{xcolor}
\usepackage{multirow}
\usepackage{booktabs}
\usepackage{float}
\usepackage{caption}
\def\BibTeX{{\rm B\kern-.05em{\sc i\kern-.025em b}\kern-.08em
    T\kern-.1667em\lower.7ex\hbox{E}\kern-.125emX}}
\begin{document}
\newcommand{\linebreakand}{%
  \end{@IEEEauthorhalign}
  \hfill\mbox{}\par
  \mbox{}\hfill\begin{@IEEEauthorhalign}
}
\newcommand{\email}[1]{\texttt{#1}}
\title{EETnet: a CNN for Gaze Detection and Tracking for Smart-Eyewear\\
\thanks{This work was carried out in the Smart Eyewear Lab, a Joint Research Center between EssilorLuxottica and Politecnico di Milano}
}

\definecolor{somegray}{rgb}{0.5, 0.5, 0.5}
\newcommand{\darkgrayed}[1]{\textcolor{somegray}{#1}}
\makeatletter
\newcommand*\titleheader[1]{\gdef\@titleheader{#1}}
\AtBeginDocument{%
  \let\st@red@title\@title
  \def\@title{%
    \vskip-1.3em
    \bgroup\normalfont\large\centering\@titleheader\par\egroup
    \vskip0.7em\st@red@title}
}
\makeatother

\titleheader{\darkgrayed{This paper has been accepted at the 
International Joint Conference on Neural Networks, 2025.
\copyright IEEE}}

 \author{\IEEEauthorblockN{Andrea Aspesi\IEEEauthorrefmark{1}\IEEEauthorrefmark{2}\IEEEauthorrefmark{3}, Andrea Simpsi\IEEEauthorrefmark{1}\IEEEauthorrefmark{2}, Aaron Tognoli\IEEEauthorrefmark{1}\IEEEauthorrefmark{2}, Simone Mentasti\IEEEauthorrefmark{2}, Luca Merigo\IEEEauthorrefmark{3}, Matteo Matteucci\IEEEauthorrefmark{2}}
 \IEEEauthorblockA{\IEEEauthorrefmark{2}Department of Electronics, Information and Bioengineering (DEIB)\\
    Politecnico di Milano - Via Ponzio 34/5, 20133 Milan, Italy\\
    \email{\{name.surname\}@polimi.it}}
 \IEEEauthorblockA{\IEEEauthorrefmark{3} EssilorLuxottica Italia S.p.A. - Piazzale Cadorna 3, 20123 Milan, Italy\\
    \email{\{name.surname\}@luxottica.com}}
}
\makeatletter

\let\@oldmaketitle\@maketitle
\renewcommand{\@maketitle}{\@oldmaketitle
    {\includegraphics[width=\linewidth,height=15\baselineskip]
    {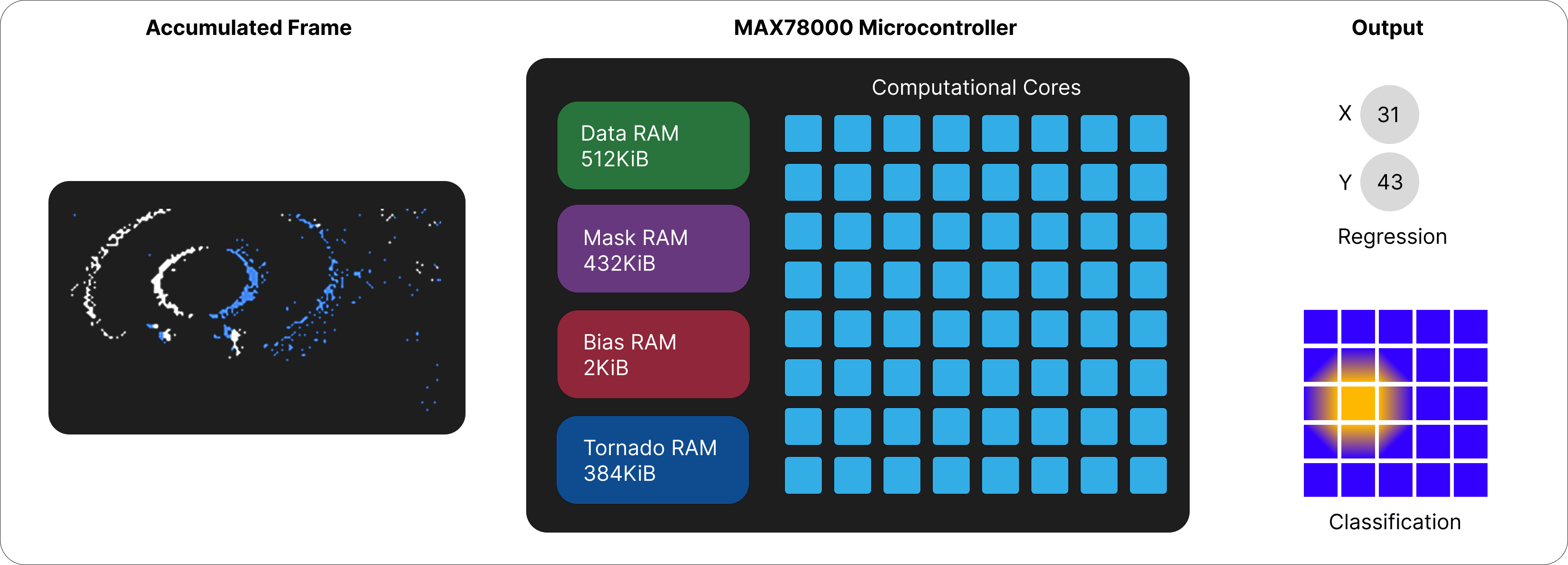}\bigskip}
    \setcounter{figure}{0}
    \captionof{figure}{Overview of the EETnet pipeline. The network, designed to run on a microcontroller, processes accumulated event frames of an eye to detect pupil location using either a regression or a classification approach.}
    \label{fig:pipeline}}
    
\makeatother

\maketitle
\begingroup\renewcommand\thefootnote{*}
\footnotetext{These authors contributed equally}
\begin{abstract}
Event-based cameras are becoming a popular solution for efficient, low-power eye tracking. Due to the sparse and asynchronous nature of event data, they require less processing power and offer latencies in the microsecond range. However, many existing solutions are limited to validation on powerful GPUs, with no deployment on real embedded devices. In this paper, we present EETnet, a convolutional neural network designed for eye tracking using purely event-based data, capable of running on microcontrollers with limited resources. Additionally, we outline a methodology to train, evaluate, and quantize the network using a public dataset. Finally, we propose two versions of the architecture: a classification model that detects the pupil on a grid superimposed on the original image, and a regression model that operates at the pixel level.
\end{abstract}

\section{Introduction} \label{introduction}
Nowadays there is a growing interest in eyes as a new machine-human interface. Quite heterogeneous applications trace eye movements to estimate parameters of the users, as focus and readiness. Eye-tracking\cite{lupu_survey_2014} is a technique that monitors eye movements to determine where a test subject is looking, what is looking at, and for how long. The resulting data gives additional insight and information on the user that can be used and elaborated in many different contexts which can vary from medical applications to entertainment ones.

In the last years, eye tracking is gaining new academic and industrial interest thanks to the rise of new advanced technologies both on the sensor side and from an algorithmic perspective.
In this context, smart eyewear has led to a growth in the availability of eye-tracking technologies~\cite{spil2019adoption}. However, these systems are not yet capable of operating in a real-time scenario or they require heavy computational power and high-end PCs to process their data~\cite{wan2021robust}. Right now, the most accurate systems are based on deep neural networks (DNNs), such as convolutional (CNNs) and recurrent (RNNs), as have been seen in the challenge~\cite{wang2024eventbasedeyetrackingais}.

Neural networks can provide high accuracy and inter-subject stability, but at the cost of computational power. 
Target platforms for such systems usually comprise GPU-powered computers, with little focus on battery-powered implementations.

Traditionally, eye tracking methods (in particular deep learning ones) leveraged cameras that can produce frame-like data (RGB, greyscale, etc.)~\cite{zafar2023investigation} that look directly to the user's eye.
In recent years, alternative solutions have been proposed, from non-camera-based such as microelectromechanical systems (MEMSs) to scan the eyes~\cite{zafar2023investigation} to photodetector-based solutions~\cite{crafa2024towards}.
Then, other approaches focused their attention on new and innovative vision sensors, in particular event-based cameras~\cite{gallego2020event}.
This new type of camera sensor detects the change of brightness at the pixel level and returns only the information of which pixel detected the change in an asynchronous stream of events. This feature can be used to detect only moving objects and, in the case of eye-tracking, to detect the pupil only when it moves. In this way, it is possible to drastically reduce the amount of data sent to the processing unit, allowing an increase in the performance of the whole system while maintaining low computational power.~\cite{zafar2023investigation}.

Exploiting only event-based data for eye-tracking is not a trivial task. Event data of the eye comprises a lot of noise, mainly from eyelids area. Machine learning approaches are a promising set of solutions. 

In this work, we present EETnet, a tiny convolutional neural network designed for eye-tracking using only event-based data that can be used in an embedded microprocessor. Figure~\ref{fig:pipeline} provides a high-level representation of the network pipeline.

This paper is organized as follows: Section~\ref{related_works} presents the current state of the art in eye-tracking, with a focus on DNN models and event-based datasets available. Section~\ref{Method} described the steps taken to develop EETnet. In Section~\ref{experimental_results}, we highlight the different experiments made with EETnet, comparing the performance of different microprocessors. Lastly, Section~\ref{conclusion} offers concluding remarks and summarizes the findings of this paper.

\section{Related Works} \label{related_works}
In this section, we discuss the different eye-tracking systems and event-based datasets used to process data coming from a sensor pointing toward the user.

The eye-tracking systems present different techniques, which range from methods that use only deep learning models to methods that employ a hybrid approach, fusing the deep learning models with the classic computer vision methods in their pipeline.

Zhao et al.~\cite{zhao_ev-eye_2023} proposed EV-Eye, a multimodal U-Net-based pipeline for pupil segmentation using near-eye grayscale images and event-based data. This method involves binarizing pupil masks, removing noise, and estimating gaze points via polynomial regression with subject-specific calibration. This method can obtain high accuracy with low latency. Unfortunately, this type of network is not suitable for use in an embedded framework.
Another hybrid approach was proposed by Poulopoulos et al.~\cite{poulopoulos_real-time_2022}, which combines a Modified Fast Radial Symmetry Transform (FRST) with a CNN to localize eye centers from full-face images. The FRST identifies probable eye center candidates, refined through a CNN for improved accuracy.

\begin{figure}[t]
    \centering
    \includegraphics[width=.48\textwidth]{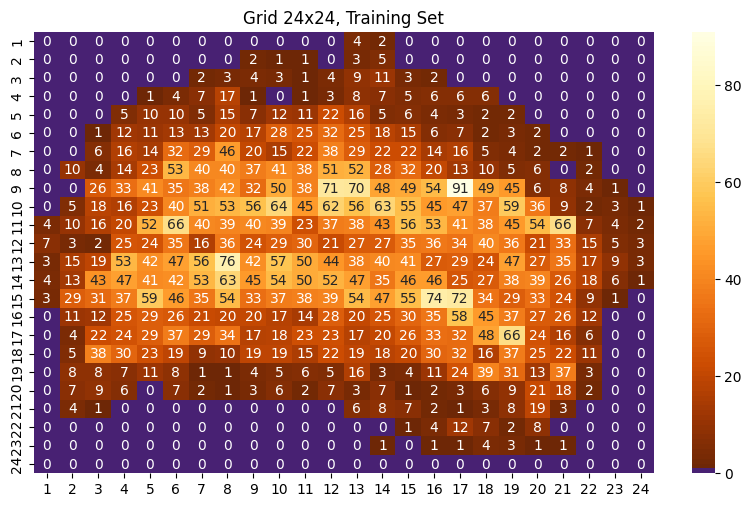}\hfill
    \includegraphics[width=.48\textwidth]{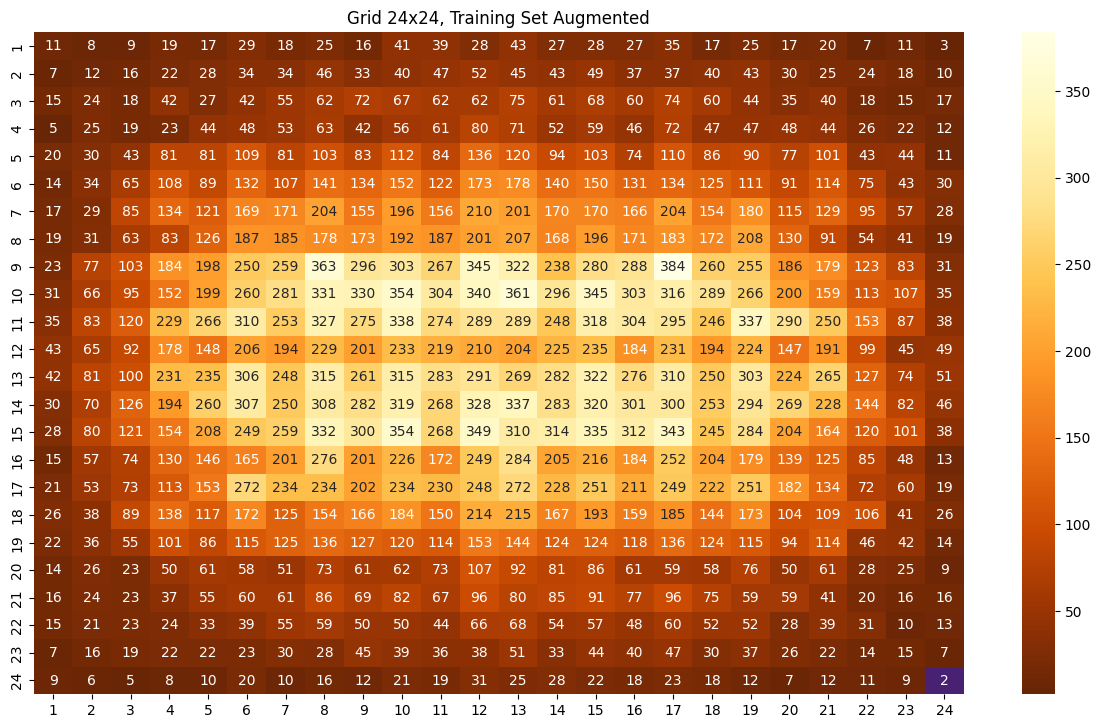}\hfill
    
    \caption[]{Heatmap of pupil center locations. The top image illustrates the original distribution, while the bottom image shows the distribution after augmentation. }
\label{fig:heatmap}    
\end{figure}

One of the earliest datasets for the eye-tracking task was introduced by Angelopoulos in~\cite{angelopoulos2020event}. This dataset was created using data collected by the DAVIS sensor, which captures both grayscale and event-based data. It includes eye movement recordings from 27 different users. However, the ground truth provided is somewhat limited due to the challenges of generating precise pupil-level annotations from the sparse event data. The dataset offers as ground truth the point of gaze on a screen rather than the pupil's position within the image frame. While this type of annotation has been widely used in the eye-tracking field for many years~\cite{holmqvist_eye_movements}, it adds complexity to the system through additional calibration requirements and nonlinearities, reducing the accuracy of evaluating pupil detection and tracking algorithms.

Zhao et al. presented a dataset as part of the EV-Eye paper~\cite{zhao_ev-eye_2023}. This dataset was acquired using two DAVIS346 sensors to obtain the event-based data and greyscale image, and a Tobii Pro Glasses 3 to obtain the gaze references. This additional metadata comprises PoGs and pupil diameters of the users at 100Hz. The dataset is composed
of 48 participants (28 male and 20 female) aged between 21 and 35 years. Labels are provided at sensor level, but only at a slow frequency since are estimated from near-eye greyscale images.

Lastly, Wang et al. have recently presented a new dataset, called 3ET+~\cite{wang2024eventbasedeyetrackingais}. This dataset has been acquired for a challenge and the data have been obtained using a DVXplorer Mini event camera. The recording consists of 13 participants doing different eye movements. 3ET+ includes ground truth annotations at 100Hz. Additionally, it provides two types of labels: a binary value indicating blink status and human-labeled coordinates of the pupil center. However, the dataset was collected without an IR-pass filter, resulting in events that include both object reflections and eye movements. This could impact the generalization capability of deep learning algorithms trained on this dataset.

\section{Method} \label{Method}
This section details the steps involved in developing EETnet, from initial data preprocessing to the final quantization required for hardware deployment.

First, we discuss the data and annotations used to train the network. EETnet processes input frames generated by accumulating events at a fixed frequency. For each frame, we apply various preprocessing and data augmentation techniques to enhance the model’s robustness and performance.

Next, we describe the architectural choices made for EETnet, balancing the need for accurate predictions with the overall efficiency of the network.

Finally, we selected an optimal quantization configuration to enable direct deployment of the network on hardware, maintaining a balance between computational efficiency and model accuracy.
\subsection{Dataset}

\begin{figure}[t]
    \centering
    \includegraphics[width=.24\textwidth]{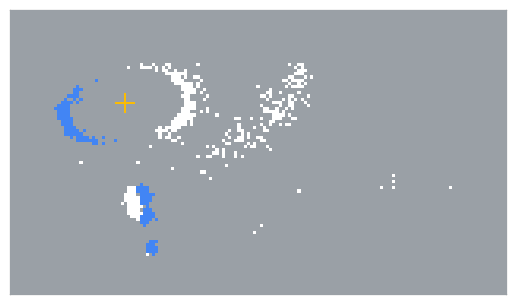}\hfill
    \includegraphics[width=.24\textwidth]{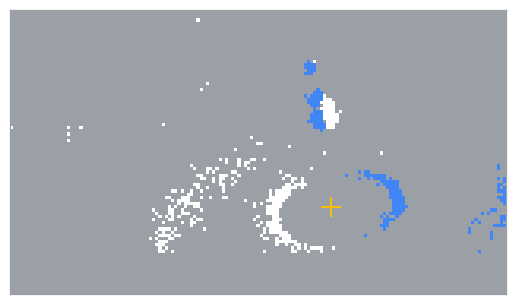}\hfill
    \includegraphics[width=.24\textwidth]{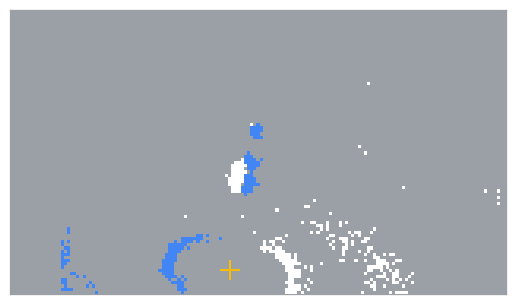}\hfill
    \includegraphics[width=.24\textwidth]{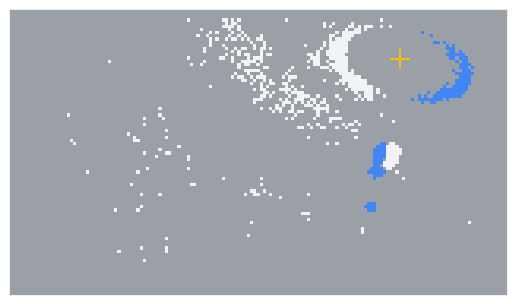}
    \caption{Example of Augmented frames. White pixels are positive events, blue are negative events, yellow cross is the labeled center.}
    \label{fig:augmentedFrame}
\end{figure}

\begin{figure*}[t]
    \centering
    \includegraphics[width=1.0\linewidth]{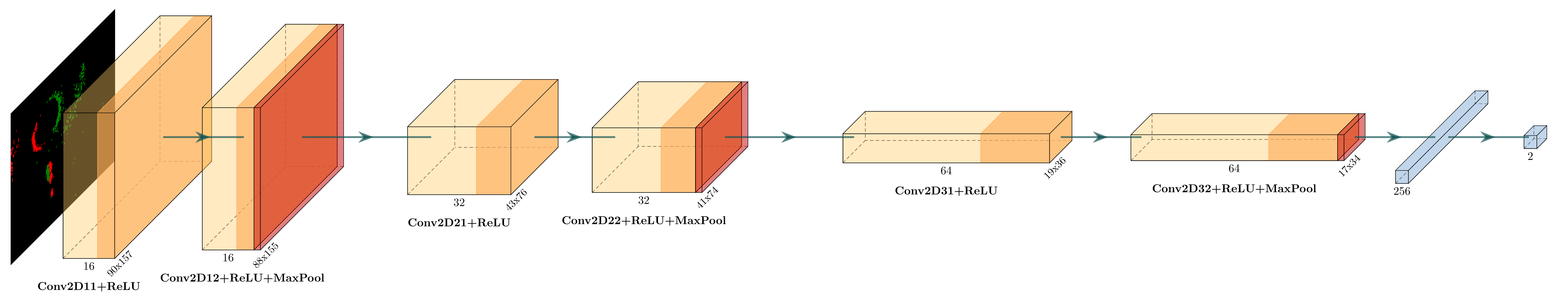}\hfill
    \includegraphics[width=1.0\linewidth]{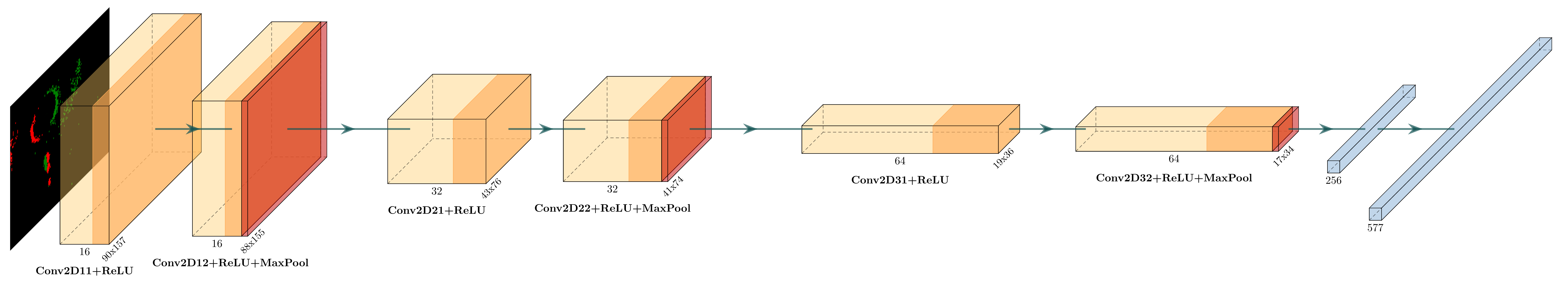}\hfill
    \caption{Schema of EETnet regression model (upper scheme) and classification model (bottom scheme), the only difference is the last layer with 577 neurons instead of 2.}
    \label{fig:EETnet}
\end{figure*}

A representative dataset is essential for training a deep learning model effectively, ensuring both accuracy and generalization. We used the previously mentioned event-based eye-tracking dataset provided by Angelopolus et al.~\cite{angelopoulos2020event}.

To generate our training set, we created event frames by accumulating events over a 5 ms window, applying a minimum threshold of 150 events per frame. This approach resulted in a 200 Hz fixed frame rate, simulating the effect of a high-speed camera.

There are different approaches in research on how to generate frames of events, other works~\cite{bonazzi2024retina} use a variable time-span instead between each frame to better compensate for the typical comet-like artifacts (similar to blurring in standard cameras) that might arise when the movement of the eye is fast. We chose a fixed interval as it's simpler to implement in hardware, and chose accordingly 200 Hz as it's faster than most eye movements~\cite{holmqvist_eye_movements}.

In order to decouple the calibration problem from the tracking application, our methodology tracks the eye directly in the camera coordinates.

However, the dataset's original annotations only included gaze positions on a screen rather than the pupil center in the camera's image frame. To address this, we annotated a substantial portion of the dataset, providing accurate ground truth data for our model.

The annotation procedure is semi-automatic, and leverages computer vision widespread methods to simplify the manual labeling procedure. A first tentative eye center is detected via match template with custom patterns, then refined via the Random Sample Consensus (RANSAC) algorithms, which provides a preliminary reference for annotation. Then, we used a custom interface to guide the labeling review process, enabling precise and verified annotation of the pupil center in each frame. A comprehensive description of the data generation and labeling procedure is provided in \cite{simpsi2025highfrequencyneareyegroundtruth}.

As the dataset was collected using a head-rest, different chin-to-eye distances of the users results in different positions of the eye movement area in the camera frames.
This is a characteristic of the user and not of the data, so a cropping and alignment procedure were implemented. The objective is to have a bounding box around the eyes at the same coordinates for every user, to have a balanced input for training the network.

The first step was then to crop all the events outside the eye region for each user, then compute the bounding box corresponding to the majority of the data. Later, an arbitrary user's bounding box center was chosen as reference, and all the other users were shifted to the common origin.
The final region of interest (ROI), and network input, was then reduced in resolution from 346x260 to 157x90 pixels. 
The resulting heatmap shown in Figure~\ref{fig:heatmap} highlights the achieved uniformity.

In addition to balancing the training data, this cropping step proves crucial to reduce the overall size of the model and allows us to better deal with the limited amount of data.

After cropping, we applied data augmentation techniques to enhance generalization and expand the training dataset.
The augmentation strategies included:
\begin{itemize}
    \item Vertical Flip (VFlip): Each frame was flipped vertically, with the objective of guiding the model not to focus on eyebrows noise.
\item Horizontal Flip (HFlip): Each frame was flipped horizontally, to generate a similar inverted movement and enhance the eye pattern relative to noise.
\item Combined Flip (VHFlip): Both vertical and horizontal flips were applied.
\item Frame Shift: Frames were shifted by a random number of pixels in the range of [-30, +30] on both axes, to add variability and better uniform coverage of the ROI.
\end{itemize}
These techniques increased the size of the dataset by a factor of eight, and in Figure~\ref{fig:augmentedFrame} you can see an example of their application.

\subsection{Network design}

The network we developed is inspired by the TinissimoYOLO architecture~\cite{moosmann_tinyissimoyolo_2023}.

The model backbone comprises three pairs of convolutional layers, each followed by ReLU activations and max pooling layers. The final layers include two fully connected layers: one processes the output from the convolutional backbone, and the other generates the final output. This final layer can be configured for either regression or classification tasks, depending on the number of neurons, as shown in Figure~\ref{fig:EETnet}.

The final architecture of the network resembles the original one as we posed ourselves the strict constraint of the target hardware, so we are limited in the possible configurations. We limited our architecture exploration on the depth of the filters and the neurons in the fully connected region, as well as removing or adding convolutional layers.

We developed two network variants, one for a regression task and another for a classification task. In the regression variant, the network outputs pixel coordinates (x, y) of the pupil center in the input frame. In the classification variant, a 24x24 grid is used to map all possible pupil positions within the input frame, plus an additional class for when the pupil is not visible, as shown in Figure 3.

To evaluate the model, we split the data into training, validation, and testing sets at the user level, ensuring that test data were sourced from users unseen during training. This ensures the model does not exploit characteristics specific to a seen user (over-fitting), but highlights generalization capability.

\subsection{Model deployment}
Deploying the model on embedded hardware, which often has limited computational power and memory, required several optimization steps to ensure efficient performance without compromising too much on accuracy.

The first optimization step was Quantization Aware Training (QAT), a technique that prepares the network for deployment at a lower precision. Typically, deep learning models use floating-point representations for weights and activations, which are computationally intensive for embedded systems, but efficient for GPU-based training. QAT mitigates this by simulating quantization effects during training, allowing the model to adapt to reduced precision, such as 8-bit integers. 

The QAT pipeline we used is based on~\cite{jacob2017quantizationtrainingneuralnetworks} and consists of four main stages:
\begin{itemize}
    \item \textit{Activation Statistics Collection}: activation statistics are collected by running the model on the training dataset.
    \item \textit{Activation Threshold Determination}: activation thresholds are determined based on the collected statistics, after an outlier removal step based on z-score.
    \item \textit{Scale Adjustments}: the scales of each layer is computed depending on the the layer's operation and connections.
    \item \textit{Weights Quantization}: while training, the model, weights and biases are fake-quantized to integers by successively performing a quantization to integer (at the specified resolution) and then de-quantized them back to floating point numbers.
\end{itemize}

These steps allow the model to fit within the resources available on the target hardware. Through QAT, the model’s weights adjust to the lower bit precision, resulting in a smaller and faster model that maintains performance. Converting weights and activations to 8-bit representations reduces memory usage by a factor of four, making the model more suitable for memory-constrained devices.

The second deployment step was generating a detailed model description file to define how the network should be executed on the target hardware. This file specified the quantized network architecture, layer count, and execution instructions for each layer on available processing units. Given the limited working memory on the embedded device, the model description file also detailed memory offsets for each layer’s output, minimizing data overlap and preventing data corruption. Where overlap was unavoidable, careful selection of offsets ensured that previous layer outputs were overwritten only after completing necessary computations.

Finally, we optimized memory usage by reusing available memory regions effectively. This was achieved by assigning specific processors to each layer to maximize memory utilization while respecting computational dependencies. This processor-layer mapping strategy allowed us to conserve memory without disrupting layer execution order.

\section{Experimental Results} \label{experimental_results}

\setlength{\tabcolsep}{2.8pt}
\begin{table}[t]
    \centering
    \footnotesize
    \begin{tabular}{lcccc}
        \toprule 
        & \textbf{FC}
        & \textbf{{\begin{tabular}[c]{@{}c@{}}Weight Size\\ (MB)\end{tabular}}}
        & \textbf{Pixel Distance}
        & \textbf{Block Distance} \\ \hline
        \vspace{1pt}
        eetnet\_256\_24\_24  &{256}     &{7.84}       &\textbf{{3.69}}        &\textbf{{3.46}} \\ \vspace{1pt}
        eetnet\_128\_24\_24  &{128}     &{4.06}       &{3.72}        &{3.49} \\ \vspace{1pt}
        eetnet\_64\_24\_24  &{64}     &\textbf{{2.17}}       &{3.89}        &{3.64} \\
        \bottomrule
    \end{tabular}
    \vspace*{3pt}
    \caption{Results of the Fully Connected models.}
    \vspace*{-10pt}
    \label{tab:FullyConnectedCutResults}
\end{table}

\subsection{Network Optimization}
The network configuration described previously was fine-tuned through a pruning process, starting with a larger initial model and iteratively reducing its size to observe the impact on performance. Here, we discuss the performance of the network under different configurations.

The first step was to reduce the network's input size even further, as this has a deep impact on the overall size of the network. Using image down-sampling, we halved the dimensions in both width and height, first starting on width only. We observed similar network accuracy, but with a reduced network size (and computational time).

We leveraged data folding techniques to reintroduce the information split into two channels. This is a direct consequence of the structure of the accelerator, where inputs less than 88x88 (and up to 4 channels) are preferred and perform better.
We observed in this scenario that this technique did not provide a real benefit over just dropping the data.

After the input, we proceeded to modify the network structure. The training routine for each tested configuration consisted of 200 epochs with a batch size of 200, using the Adam optimizer. The dataset was split at the user level to ensure that test data came from users not encountered during training. Specifically, the training set included data from three users (10, 18, and 20), the validation set included one user (19), and the test set consisted of two users (5, 15). We tested three configurations with varying neuron counts in the first fully connected layer: 256, 128, and 64 neurons, respectively.

As shown in Table~\ref{tab:FullyConnectedCutResults}, results indicate that even with fewer neurons, the network maintained similar performance levels. This suggests that a more compact model architecture can achieve efficient performance with a reduced computational load and memory usage. The configuration selected to respect the memory constraints of the accelerator was 64 neurons.

\setlength{\tabcolsep}{2.8pt}
\begin{table}[t]
    \centering
    \footnotesize
    \begin{tabular}{lcccccc}
        \toprule 
        & \textbf{Type}
        & \textbf{{\begin{tabular}[c]{@{}c@{}}Kernel\\ (uJ/ms)\end{tabular}}}
        & \textbf{{\begin{tabular}[c]{@{}c@{}}Input\\ (uJ/ms)\end{tabular}}}
        & \textbf{{\begin{tabular}[c]{@{}c@{}}Input + Inference\\ (uJ/ms)\end{tabular}}}
        & \textbf{Unit} \\ \hline                  
        \multirow{4}{*}{EETnetR4}     
                              \vspace{1pt}
                              & \textbf{E} & 57    & 2.0   & 64.0  & uJ   \\
                              \vspace{1pt}
                              & \textbf{T} & 5.9   & 0.227 & 3.0   & ms   \\
                              \vspace{1pt}
                              & \textbf{I} & 8.6   & 8.6   & 8.6   & mW   \\
                              & \textbf{A} & 18.4  & 17.4  & 29.4  & mW   \\ \cline{2-6}
        \multirow{4}{*}{EETnetR8}
                              \vspace{1pt}
                              & \textbf{E} & 79    & 2.0   & 73.0  & uJ   \\ 
                              \vspace{1pt}
                              & \textbf{T} & 8.3   & 0.227 & 3.0   & ms   \\ 
                              \vspace{1pt}
                              & \textbf{I} & 9.0   & 9.0   & 9.0   & mW   \\ 
                              & \textbf{A} & 18.5  & 17.4  & 33.1  & mW   \\ \cline{2-6}
        \multirow{4}{*}{EETnetRAll4}
                              \vspace{1pt}
                              & \textbf{E} & 38    & 2.0   & 62.0  & uJ   \\
                              \vspace{1pt}
                              & \textbf{T} & 4.2   & 0.227 & 3.0   & ms   \\ 
                              \vspace{1pt}
                              & \textbf{I} & 8.7   & 8.7   & 8.7   & mW   \\ 
                              & \textbf{A} & 17.8  & 17.1  & 28.9  & mW   \\ \cline{2-6}
        \multirow{4}{*}{EETnetR2248}
                              \vspace{1pt}
                              & \textbf{E} & 56    & 2.0   & 57.0  & uJ   \\ 
                              \vspace{1pt}
                              & \textbf{T} & 5.8   & 0.227 & 3.0   & ms   \\ 
                              \vspace{1pt}
                              & \textbf{I} & 8.7   & 8.7   & 8.7   & mW   \\ 
                              & \textbf{A} & 18.2  & 17.5  & 27.3  & mW   \\ \cline{2-6}
                              
        \multirow{4}{*}{EETnetR1248}
                              \vspace{1pt}
                              & \textbf{E} & 54    & 2.0   & 52.0 & uJ   \\
                              \vspace{1pt}
                              & \textbf{T} & 5.8   & 0.227 & 3.0   & ms   \\ 
                              \vspace{1pt}
                              & \textbf{I} & 9.0   & 9.0   & 9.0   & mW   \\ 
                              & \textbf{A} & 18.3  & 17.5  & 26.0  & mW   \\ 
        \bottomrule
    \end{tabular}
    \vspace*{3pt}
    \caption{Energy and time measurements for EETnet regression models. Legend: \textbf{E} = Energy transformed during the window, \textbf{T} = Active measurement period, \textbf{I} = Idle period power measurement, \textbf{A} = Active period power measurement.}
    \vspace*{-10pt}
    \label{tab:energyMeasurements}
\end{table}

\setlength{\tabcolsep}{2.8pt}
\begin{table}[t]
    \centering
    \footnotesize
    \begin{tabular}{lccc}
        \toprule 
        & \textbf{Metric}
        & \textbf{Value}
        & \textbf{Unit} \\ \hline
                  
        \multirow{4}{*}{EETnetR4}
            \vspace{1pt}
            & Mean Absolute Error       &\textbf{2.55}            & px      \\ \vspace{1pt}
            & Mean Pixel Distance       &\textbf{4.05}            & px      \\
            \vspace{1pt}
            & Mean Angle Error          & \textbf{3.09}   & °       \\
            \vspace{1pt}
            & Weights Size              & 84.45           & kB      \\ 
            \cline{2-4}
        \multirow{4}{*}{EETnetR8}
            \vspace{1pt}
            & Mean Absolute Error       & 3.21            & px      \\
            \vspace{1pt}
            & Mean Pixel Distance       & 5.12            & px      \\
            \vspace{1pt}
            & Mean Angle Error          & 3.80            & °       \\
            \vspace{1pt}
            & Weights Size              & 119.62          & kB      \\ \cline{2-4}
        \multirow{4}{*}{EETnetRAll4}
            \vspace{1pt}
            & Mean Absolute Error       & 4.42            & px      \\
            \vspace{1pt}
            & Mean Pixel Distance       & 6.95            & px      \\
            \vspace{1pt}
            & Mean Angle Error          & 5.26            & °       \\
            \vspace{1pt}
            & Weights Size              & \textbf{59.81}  & kB      \\ \cline{2-4}
        \multirow{4}{*}{EETnetR2248}  
            \vspace{1pt}
            & Mean Absolute Error       & 2.84            & px      \\
            \vspace{1pt}
            & Mean Pixel Distance       & 4.50            & px      \\ 
            \vspace{1pt}
            & Mean Angle Error          & 3.44            & °       \\
            \vspace{1pt}
            & Weights Size              & 84.15           & kB      \\ \cline{2-4}
        \multirow{4}{*}{EETnetR1248}
            \vspace{1pt}
            & Mean Absolute Error       & 5.15            & px      \\
            \vspace{1pt}
            & Mean Pixel Distance       & 8.19            & px      \\
            \vspace{1pt}
            & Mean Angle Error          & 6.06            & °       \\
            \vspace{1pt}
            & Weights Size              & 84.14           & kB      \\
        \bottomrule
    \end{tabular}
    \vspace*{3pt}
    \caption{Performance metrics for EETnet regression models.}
    \vspace*{-10pt}
    \label{tab:performanceMetrics}
\end{table}

\subsection{Network Performance on Hardware}
\begin{figure*}[t]
    \centering
    \includegraphics[width=0.85\textwidth]{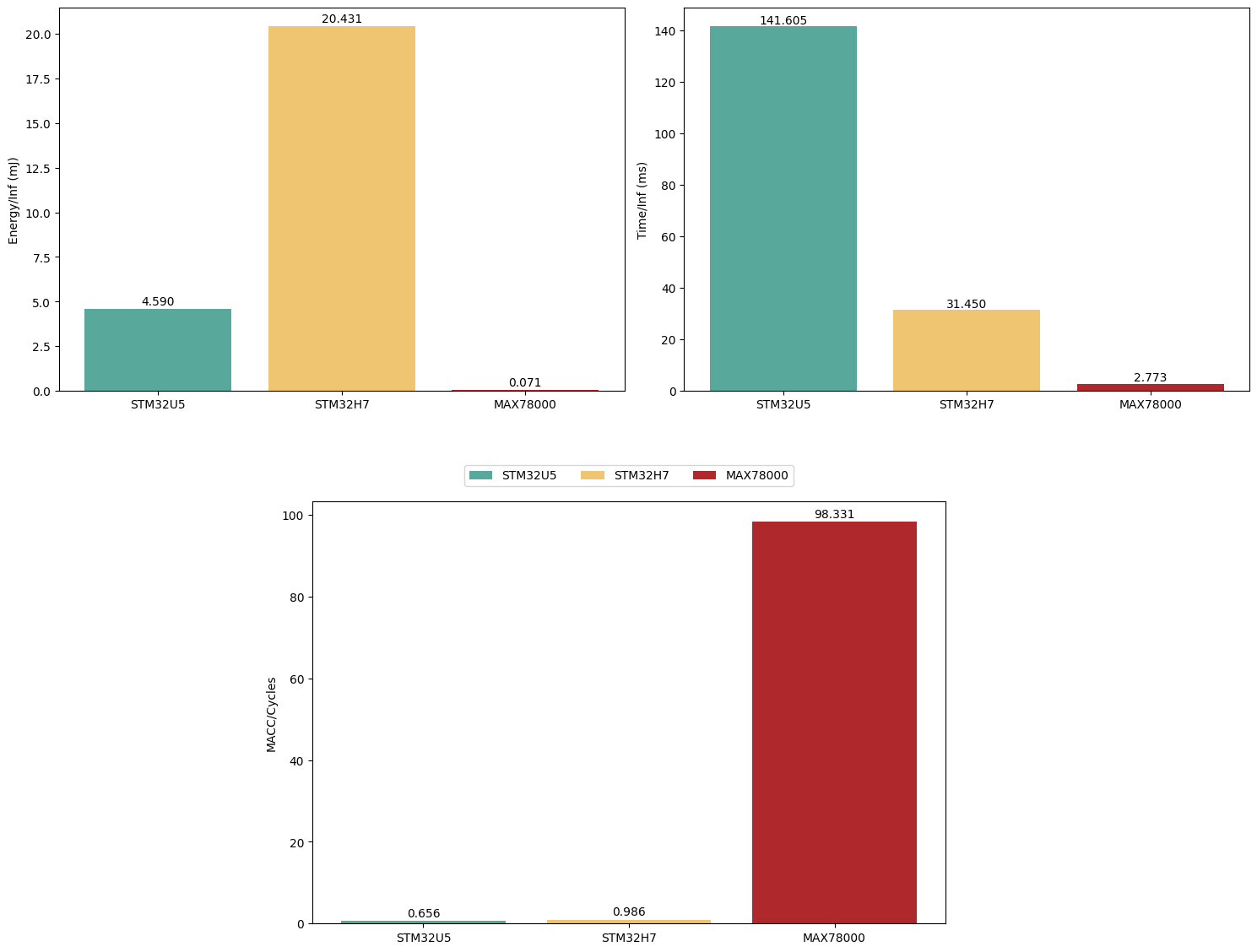}
    \caption{Comparisons of energy consumption per inference, time per inference, and Multiply and Accumulate operations over cycles on 3 different microcontrollers: STM32U5, STM32H7, and MAX78000.}
    \label{fig:energyComparisons}
\end{figure*}

After selecting the final architecture, we tested the best quantization configuration. The following configurations were evaluated: 
\begin{itemize}
    \item EETnetR4: Regression configuration, the weights of the convolutional layers are quantized using 4 bits, instead 8 are used for the fully connected ones. 
    \item EETnetR8: Regression configuration, all the layers have been quantized using 8 bits.
    \item EETnetRAll4: Regression configuration, all the layers have been quantized using 4 bits.
    \item EETnetR2248: Regression configuration, the weights of the first two convolutional layers are quantized using 2, for the remaining convolutional ones are quantized using 4 bits, and 8 bits are used for the fully connected ones.
    \item EETnetR1248: Regression configuration, the weights of the first convolutional layer are quantized using 1 bit, 2 bits are used for the second convolutional layer, 4 bits for the remaining convolutional layers, and 8 bits are used for the fully connected ones.

    \item EETnetC4: Classification configuration, the weights of the convolutional layers are quantized using 4 bits, instead 8 are used for the fully connected ones.
    \item EETnetC8: Classification configuration, all the layers have been quantized using 8 bits.
    \item EETnetCAll4: Classification configuration, all the layers have been quantized using 4 bits.
    \item EETnetC2248: Classification configuration, the weights of the first two convolutional layers are quantized using 2, for the remaining convolutional ones are quantized using 4 bits, and 8 bits are used for the fully connected ones.
    \item EETnetC1248: Classification configuration, the weights of the first convolutional layer are quantized using 1 bit, 2 bits are used for the second convolutional layer, 4 bits for the remaining convolutional layers, and 8 bits are used for the fully connected ones.
\end{itemize}
Regression configuration is when the network has 2 neurons for the last layer, and classification configuration has 577 neurons in the last layer.

The results, summarized in Tables \ref{tab:performanceMetrics} and \ref{tab:performanceMetricsC}, show that the optimized versions of EETnet not only meet performance requirements but may offer slight accuracy improvements, potentially due to enhanced generalization. These optimized models also have lower power consumption, as seen in Tables \ref{tab:energyMeasurements} and \ref{tab:energyMeasurementsC}, and reduced memory footprints for weights, making them ideal for energy-efficient, resource-limited applications. These values are in line with requirements for battery-powered devices, paving the way for integration into wearable or portable systems.

Both the classification and regression variants of EETnet were successfully flashed and tested on sample data onto the accelerator. The network’s behavior mirrored the results obtained during emulated testing on the development machine, underscoring the reliability of the deployment process. The quantized models exhibited performance consistent with the baseline metrics, maintaining accuracy in eye-tracking tasks. EETnet is then capable of operating in real-time scenarios on embedded hardware, meeting the demands of low-latency applications.

\subsection{Evaluation on other hardware} 
The regression model is more suitable for real-life deployment as the output of the network is faster to be accessed. This is particularly true when we aim at high accuracy, as the equivalent classification problem will require a quadratic number of classes with an increase of grid dimensions.

We then deployed the network on widespread micro-controllers as comparison, to highlight the trade-off between flexility of network structure and resulting performance and power consumption.

General purpose platform cannot leverage the benefits of extreme quantization (less than 8-bit integers), so here we use the EETnetR8 model. The comparison has been made between: STM32U5, STM32H7, and MAX7800.
As it's possible to see in Figure~\ref{fig:energyComparisons}, the MAX78000 achieves much better performance compared to the other two microcontrollers. 
Energy-wise the MAX78000 is the only one that achieves a consumption of less than 1mJ  per inference, accompanied by an inference time of less than 3ms, while the other two are far over 10ms.
Moreover, due to its parallel processors, the MAX78000 can perform much more than one MACC operation per clock cycle. The MAX78000 can achieve these results since it's tailored specifically to execute CNN layers quickly and efficiently. This performance comes at the cost of a limited instruction set and an ad hoc training and deployment procedure.

\setlength{\tabcolsep}{2.8pt}
\begin{table}[t]
    \centering
    \footnotesize
    \begin{tabular}{lcccc}
        \toprule 
        & \textbf{{\begin{tabular}[c]{@{}c@{}}Energy\\ (Kernel)\end{tabular}}}
        & \textbf{{\begin{tabular}[c]{@{}c@{}}Energy\\ (Input)\end{tabular}}}
        & \textbf{{\begin{tabular}[c]{@{}c@{}}Energy\\ (Input + Inference)\end{tabular}}}
        & \textbf{Unit} \\ \hline
                  \vspace{1pt}
                  EETnetC4      & 81    & 2     & 64            & uJ    \\
                  \vspace{1pt}
                  EETnetC8      & 107   & 2     & 76            & uJ    \\
                  \vspace{1pt}
                  EETnetC8      & 107   & 2     & 76            & uJ    \\
                  \vspace{1pt}
                  EETnetCAll4   & \textbf{51}    & 2     & 63            & uJ    \\
                  \vspace{1pt}
                  EETnetC2248   & 82    & 2     & 58            & uJ    \\
                  \vspace{1pt}
                  EETnetC1248   & 80    & 2     &\textbf{55}    & uJ    \\  
        \bottomrule
    \end{tabular}
    \vspace*{3pt}
    \caption{Energy measurements for EETnet classification models.}
    \vspace*{-10pt}
    \label{tab:energyMeasurementsC}
\end{table}

\setlength{\tabcolsep}{2.8pt}
\begin{table}[t]
    \centering
    \footnotesize
    \begin{tabular}{lccc}
        \toprule 
        & \textbf{Performance Metric}
        & \textbf{Value}
        & \textbf{Unit} \\ \hline
        \vspace{1pt}
        \multirow{4}{*}{EETnetC4} 
            & Mean Block Distance       & 3.47           & px    \\
            & Mean Pixel Distance       & 3.69           & px            \\
            & Mean Angle Error          & \textbf{2.87}  & °             \\
            & Weights Size              & 121.25         & kB            \\ \cline{2-4}
        \vspace{1pt}
        \multirow{4}{*}{EETnetC8}
            & Mean Block Distance       & \textbf{3.46}           & px            \\
            & Mean Pixel Distance       & \textbf{3.67}           & px            \\ 
            & Mean Angle Error          & 2.91           & °             \\ 
            & Weights Size              & 156.42         & kB            \\ \cline{2-4}
        \vspace{1pt}
        \multirow{4}{*}{EETnetCAll4} 
            & Mean Block Distance    & 3.53           & px              \\ 
            & Mean Pixel Distance       & 3.76           & px           \\ 
            & Mean Angle Error          & 2.94           & °            \\ 
            & Weights Size              & \textbf{78.21} & kB           \\ \cline{2-4}
        \vspace{1pt}
        \multirow{4}{*}{EETnetC2248}
            & Mean Block Distance    & 3.82           & px               \\ 
            & Mean Pixel Distance       & 4.07           & px            \\ 
            & Mean Angle Error          & 3.18           & °             \\ 
            & Weights Size              & 120.95         & kB            \\ \cline{2-4}
        \vspace{1pt}
        \multirow{4}{*}{EETnetC1248}
            & Mean Block Distance    & 6.99           & px            \\ 
            & Mean Pixel Distance       & 7.47           & px            \\ 
            & Mean Angle Error          & 5.62           & °             \\ 
            & Weights Size              & 120.94         & kB            \\
        \bottomrule
    \end{tabular}
    \vspace*{3pt}
    \caption{Performance metrics for EETnet classification models}
    \vspace*{-10pt}
    \label{tab:performanceMetricsC}
\end{table}

\section{Conclusion} \label{conclusion}
With this work, we presented EETnet, an eye-tracking system based on a CNN that can be used in an embedded framework. With EETnet, we demonstrate the feasibility and efficacy of a network that can run on a microprocessor, achieving low energy consumption and operating solely on event-based data.
The results highlight that, while the network focuses on computational efficiency and low inference time, it achieves a good level of accuracy, making it a viable solution for real-world embedded applications.

\bibliographystyle{splncs04}
\bibliography{bibliography}
\end{document}